\documentclass[sigconf,authorversion,screen]{acmart} %

\AtBeginDocument{%
  \providecommand\BibTeX{{%
    \normalfont B\kern-0.5em{\scshape i\kern-0.25em b}\kern-0.8em\TeX}}}

\copyrightyear{2023}
\acmYear{2023}
\setcopyright{acmlicensed}\acmConference[KDD '23]{Proceedings of the 29th ACM SIGKDD Conference on Knowledge Discovery and Data Mining}{August 6--10, 2023}{Long Beach, CA, USA}
\acmBooktitle{Proceedings of the 29th ACM SIGKDD Conference on Knowledge Discovery and Data Mining (KDD '23), August 6--10, 2023, Long Beach, CA, USA}
\acmPrice{15.00}
\acmDOI{10.1145/3580305.3599520}
\acmISBN{979-8-4007-0103-0/23/08}

\usepackage{algorithm}
\usepackage{algorithmicx}
\usepackage{algpseudocodex}
\usepackage{xfrac}
\usepackage{balance}

\graphicspath{{figures/}}

\begin{document}

\newcommand{\textbs}[1]{\textcolor{red}{\textbf{#1}}}
\newcommand{\textgd}[1]{\textcolor{violet}{#1}}

\title{The Information Pathways Hypothesis: Transformers are Dynamic Self-Ensembles}

\author{Md Shamim Hussain}
\email{hussam4@rpi.edu}
\orcid{0000-0002-0832-913X}
\affiliation{%
  \institution{Rensselaer Polytechnic Institute}
  \city{Troy}
  \state{New York}
  \country{USA}
}

\author{Mohammed J. Zaki}
\email{zaki@cs.rpi.edu}
\orcid{0000-0003-4711-0234}
\affiliation{%
  \institution{Rensselaer Polytechnic Institute}
  \city{Troy}
  \state{New York}
  \country{USA}
}

\author{Dharmashankar Subramanian}
\email{dharmash@us.ibm.com}
\orcid{0000-0002-1990-7740}
\affiliation{%
  \institution{IBM T. J. Watson Research Center}
  \city{Yorktown Heights}
  \state{New York}
  \country{USA}
}
\renewcommand{\shortauthors}{Md Shamim Hussain, Mohammed J. Zaki and Dharmashankar Subramanian}

\begin{abstract}
  Transformers use the dense self-attention mechanism which gives a lot of flexibility for long-range connectivity. Over multiple layers of a deep transformer, the number of possible connectivity patterns increases exponentially. However, very few of these contribute to the performance of the network, and even fewer are essential. We hypothesize that there are sparsely connected sub-networks within a transformer, called information pathways which can be trained independently. However, the dynamic (i.e., input-dependent) nature of these pathways makes it difficult to prune dense self-attention during training. But the overall distribution of these pathways is often predictable. We take advantage of this fact to propose Stochastically Subsampled self-Attention (SSA) -- a general-purpose training strategy for transformers that can reduce both the memory and computational cost of self-attention by 4 to 8 times during training while also serving as a regularization method -- improving generalization over dense training. We show that an ensemble of sub-models can be formed from the subsampled pathways within a network, which can achieve better performance than its densely attended counterpart. We perform experiments on a variety of NLP, computer vision and graph learning tasks in both generative and discriminative settings to provide empirical evidence for our claims and show the effectiveness of the proposed method.
\end{abstract}



\maketitle
\section{Introduction}
\begin{figure}[!t]
  \centering
  \includegraphics[width=0.81\columnwidth]{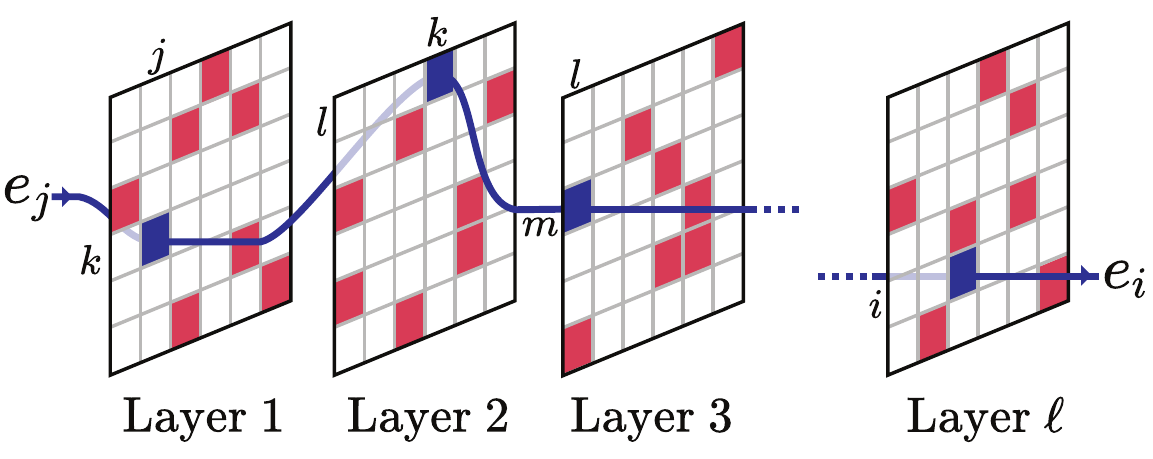}
  \caption{\small A communication channel from element $j$ to element $i$ that spans multiple layers. $e_i$ is the embedding of element $i$.}
  \Description{A conceptual diagram showing a communication channel between two elements of the input that spans multiple layers.}
  \label{fig:channel}
\end{figure}

Transformer neural networks \citep{vaswani2017attention} have become ubiquitous in all fields of machine learning including natural language processing (NLP) \citep{devlin2018bert,radford2018improving}, computer vision \citep{dosovitskiy2020image,liu2021swin}, and graph learning \citep{ying2021transformers,hussain2022global}. The transformer architecture is based on the attention mechanism \citep{bahdanau2014neural}, which allows the model to learn to focus on the most relevant parts of the input. The global self-attention mechanism allows the transformer to update the representation of each element (e.g., token, pixel, node) of the input based on that of all other elements. The relevancy of each element is dictated by the attention weights formed by the network during the update and can be expressed as the self-attention matrix. These weights are dynamically computed by the network for each particular input. This form of flexible weighted aggregation is the key to the success of the transformer. However, the all-to-all nature of the self-attention process incurs a compute and memory cost that increases quadratically with the number of input elements $N$. Consequently, the self-attention process is the main efficiency bottleneck when the transformer is applied to long inputs. During the self-attention process, if element $i$ applies a significant weight to element $j$, information can flow from $j$ to $i$ allowing them to communicate. This way, the self-attention process allows inter-element connections to form arbitrarily within a layer. However, as shown in Fig.~\ref{fig:channel}, in a deep network, this communication may occur indirectly over multiple layers, for example, element $k$ may get updated from element $j$ and then element $i$ may get updated from element $k$ in the next layer, forming a communication channel that spans multiple layers. Over $l$ layers, thus there are at least $N^{l-1}$ possible ways for the two elements to communicate. The question that arises is whether all of these exponential numbers of connections contribute to the performance of the network and if not whether some of them can be pruned to save memory and computation costs during training.

Previous works like \citep{liu2021transformer} have shown that the attention matrices of a fully trained transformer are sparse, and a large portion of its elements can be pruned without hurting inference time performance. Despite this sparsity, over multiple layers, connectivities can reach most elements of the input, similar to expander graphs. This inspired some works to pre-define a fixed sparsity pattern to the self-attention matrix \citep{child2019generating,zaheer2020big}. However, this comes at the cost of expressivity since the model is forced to learn the attention weights within the specified \emph{fixed} sparsity pattern. While the underlying connectivity in the self-attention process is sparse, this pattern is also \emph{dynamic}, i.e., input-dependent and should not be pre-imposed. Also, these connectivities do not work in isolation within a layer but expand over multiple layers to form directed subgraphs of connectivity patterns. We call these dynamically formed sparsely connected subnetworks within the fully connected transformer \emph{information pathways}. We hypothesize that not only do these pathways use a small portion of the self-attention matrix at each layer to make connections, but there are many such pathways within the network which can work independently. An ensemble of sub-models formed from a subset of pathways can often get performance close to that of the full model. Thus, we hypothesize that the transformer can be viewed as an ensemble of these sub-models, which are internally aggregated by the attention process. We use the term \emph{self-ensemble} to point out that all of these sub-models use the same set of transformer weights, and vary only in inter-element connectivity. These connectivities are input dependent, and the transformer uses the pathways to perform dynamic inference on each element of the input based on the other elements. We call the information pathways that contribute to the generalization performance of the transformer \emph{important pathways}, while other pathways can be deemed redundant or may even overfit the training data. To train a transformer, it is enough to ensure that these important pathways get enough training.

Previously, there has been a wealth of research on pruning the learnable weights of a neural network \citep{lecun1989optimal,hassibi1992second,han2015learning,li2016pruning} which reduces the cost of inference. The lottery ticket hypothesis by \citet{frankle2018lottery} states that such pruning is possible because of the existence of winning tickets -- very sparse subnetworks that exist within the dense network, as early as the initialization. When trained in isolation, these winning tickets can match or even exceed the performance of the dense network. Our information pathways hypothesis makes similar statements about the interconnectivity of the input elements and the dynamic weights of the attention matrix. Similar to the learnable weights, at inference time, the self-attention matrix can be dynamically pruned to reduce the inference cost both in terms of memory and compute \citep{qu2022dota,chen2022dynamic}. However, this is much trickier during training since the weights of the network are updated in each training step and the pruning pattern is harder to predict. In other words, unlike the winning tickets in the lottery ticket hypothesis, the important information pathways are dynamic, changing from one training sample to another. However, the connectivity patterns of the information pathways can often follow a predictable distribution. We can thus perform biased subsampling to increase the probability of covering important pathways during training while reducing the cost of training.

Our contributions are as follows -- we propose a novel method for training transformers called \textbf{SSA} (Stochastically Subsampled self-Attention) that reduces both the memory and computational requirements of training while also improving generalization. SSA works by randomly subsampling the self-attention process at each training step, which allows the model to learn different connectivity patterns. We can utilize the locality of connectivity (the local inductive bias) to perform a more intelligent subsampling than random subsampling. We show that SSA can also be performed at inference time to build a self-ensemble of sub-models, each containing a subset of pathways, which can further improve generalization. We propose the information pathways hypothesis as an implication of our empirical results, which states the existence of a small number of sparsely connected and dynamic subnetworks within the transformer, the information pathways, that can be trained independently.

\section{Related Work}
Randomly dropping part of the network such as activations \citep{srivastava2014dropout}, weights \citep{wan2013regularization} or layers \citep{huang2016deep} have been seen to improve generalization. For transformers, similarly, dropping attention weights \citep{zehui2019dropattention} and attention heads \citep{zhou2020scheduled} have led to better generalization. Among these methods, only a few such as \citep{huang2016deep} lead to a reduction in training costs. Although dropout was originally formulated for the learnable weights of a network, they were directly adopted for the attention weights \citep{zehui2019dropattention}, which empirically improves generalization. We believe that attention dropout also trains an ensemble of pathways through the network. However, unlike attention dropout, we perform subsampling in a structured manner so that we may save training costs. We also apply local inductive bias while doing so.

After training, pruning parts of the transformer can lead to a reduction in the number of parameters and save memory \citep{prasanna2020bert,chen2020lottery}, and can potentially improve generalization \citep{michel2019sixteen} and/or efficiency \citep{fan2019reducing} during inference. Our method is focused on stochastically dropping parts of the attention mechanism during training to reduce training costs, and can be used alongside the aforementioned methods. Additionally, we show the regularization effect of SSA and better generalization through ensembles of sparsely connected sub-models during inference.

Our method can also facilitate training on longer inputs, due to the reduction in both the memory and compute cost of self-attention. Previously, many works sought to remedy the computational bottleneck of dense self-attention via architectural modifications. This includes the use of sparse or localized self-attention \citep{child2019generating,zaheer2020big,beltagy2020longformer,liu2021swin}, or low-rank/linear/factorized attention \citep{choromanski2020rethinking,schlag2021linear,katharopoulos2020transformers,wang2020linformer} or recurrence \citep{dai2019transformer,rae2019compressive} and other methods \citep{kitaev2020reformer,xiong2021nystromformer}. These often make trade-offs in terms of expressivity, performance or generality to gain efficiency. Recently, many specialized architectures have evolved \citep{roy2021efficient,khandelwal2019generalization,ren2021combiner}. Despite these innovations, simple dense and local window based attention mechanisms remain relevant and competitive in many applications \citep{xiong2021simple}. Unlike these approaches, we make innovations in training transformers while allowing fall-back to vanilla dense or locally dense attention at inference time.

Many innovations have also been made to reduce the training cost of transformers on long sequences. Shortformer \citep{press2020shortformer} uses a staged training scheme where training is done first on short inputs followed by longer input sequences, which reduces the cost of training. Curriculum learning has also been used to stabilize training and optimize for large batches \citep{li2021curriculum}. However, these approaches have only been effective in causal generative language modeling or non-causal masked language modeling tasks. Our SSA is applicable to any causal/non-causal generative or discriminative tasks, on any form of data including text, images, and graphs.

Our self-ensembling method is related to the ensemble methods of neural networks \citep{hansen1990neural,huang2017snapshot,izmailov2018averaging}. However, unlike these methods, we do \emph{not} train multiple models and average their predictions/weights. Instead, we train a single model with SSA and form an ensemble of sub-models at inference time using different subsampled attention patterns. This approach resembles Monte Carlo dropout \citep{gal2016dropout}, which performs dropout at inference time to make multiple predictions for uncertainty estimation. However, while MC dropout randomly drops activations, we subsample the attention mechanism from a specific distribution. Our main focus is improving generalization through self-ensembling, while its potential use for uncertainty estimation is left for future work.

\section{Method}
\subsection{Background}
The transformer architecture \citep{vaswani2017attention} consists of an encoder and a decoder. An encoder-only architecture can be used for tasks like classification \citep{dosovitskiy2020image} and masked language modeling \citep{devlin2018bert}, whereas a decoder-only architecture can be used for generative tasks \citep{radford2018improving,chen2020generative}. Both of these only require self-attention. For tasks like machine translation, an encoder-decoder architecture is used which additionally uses cross-attention in the decoder. We only focus on the self-attention mechanism of the transformer in this work.
The key innovation of the transformer is the multihead attention mechanism, which can be expressed as:
\begin{align}
  \mathrm{Attn}(\mathbf{Q}, \mathbf{K}, \mathbf{V}) = \mathrm{softmax}\left(\frac{\mathbf{Q} \mathbf{K}^T}{\sqrt{d_k}}\right)\mathbf{V} \label{eq:xformer_enc}  =\mathbf{A}\mathbf{V}
\end{align}
where $\mathbf{Q}, \mathbf{K}, \mathbf{V}$ are matrices containing rows of keys, queries and values. In the case of self-attention, all of them are formed by learnable projections of the embeddings. $d_k$ is the dimensionality of the queries and the keys. $\mathbf{A}$ is known as the attention matrix. Element $(i,j)$ of this matrix is formed from the scaled dot product of query $q_i$ and the key $k_j$ followed by a softmax over all $j$. The normalized weights at row $i$ are used to aggregate the values $v_j$ in updating the representation of position $i$, thus allowing information to flow from $j$ to $i$. This process is done for multiple sets of queries, keys and values, where each is called an \emph{attention head}.

Several other terms may be added to the scaled dot product of queries and keys. A masking value $m_{ij}=-\infty$ may be added to prevent the model from attending to future positions (i.e., $j>i$) for generative modeling or to padding tokens; the softmax function drives the attention matrix to zero at these positions. Another term may be added to encode relative positions. Although this may take different forms \citep{shaw2018self,dai2019transformer,roberts2019exploring,press2021train,wu2021rethinking,park2022grpe}, we will discuss methods where a relative positional bias $r_{i-j}$ is added to the scaled dot-product, e.g., \citep{roberts2019exploring,liu2021swin,press2021train}. Our method should apply to other forms of relative positional encodings as well. With the inclusion of masking and relative positional encodings, the attention matrix becomes:
\begin{align}
  \mathbf{A} = \mathrm{softmax}\left(\frac{\mathbf{Q} \mathbf{K}^T}{\sqrt{d_k}}+\mathbf{M}+\mathbf{R}\right)
  = \mathrm{softmax}\left(\frac{\mathbf{Q} \mathbf{K}^T}{\sqrt{d_k}}+\mathbf{B}\right) \label{eq:xformer_final} 
\end{align}
Where, $\mathbf{M}$ is the masking matrix and $\mathbf{R}$ is the relative positional bias matrix. We merge both of these into a single bias matrix $\mathbf{B}$.

\begin{figure}[!t]
  \centering
  \includegraphics[width=1.0\columnwidth]{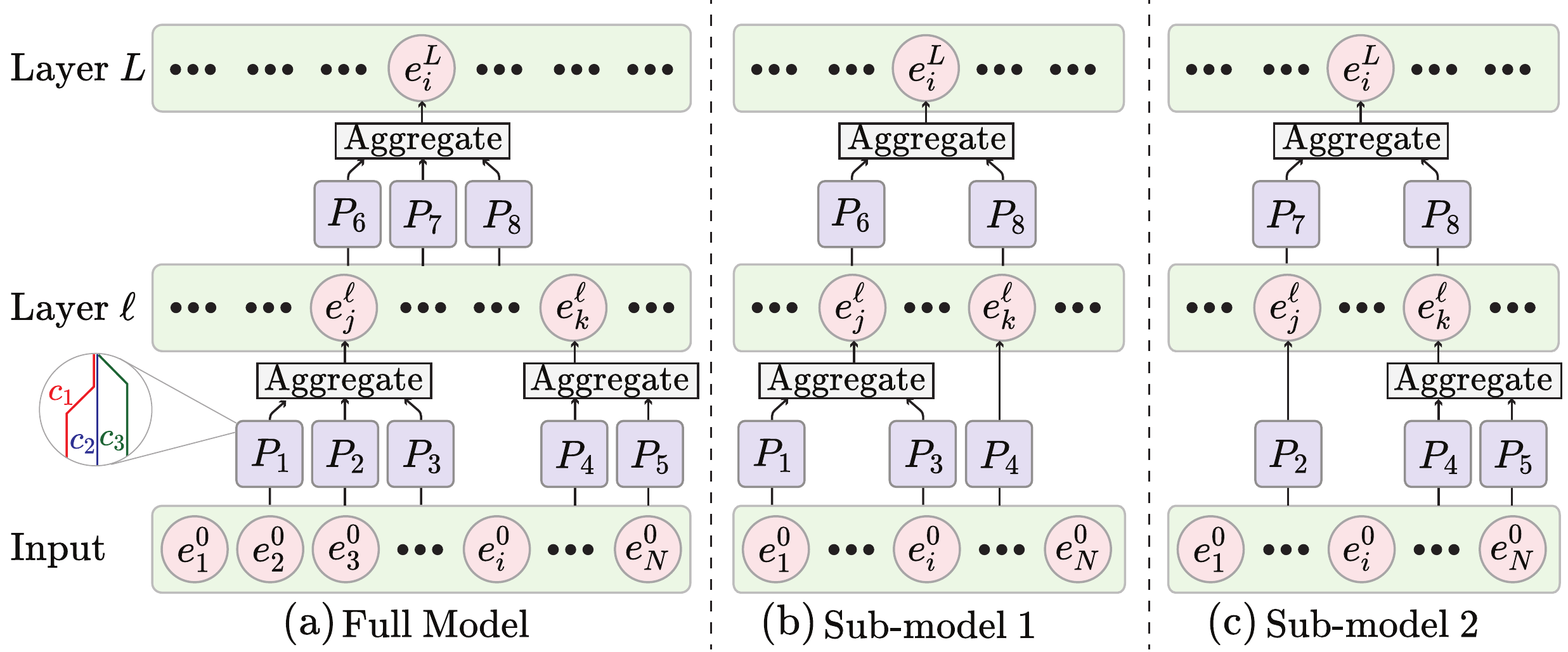}
  \caption{\small A conceptual demonstration of the information pathways hypothesis. Embeddings are $e_i$, information pathways are $P_i$, and communication channels are $c_i$. (a) is the full model, (b) and (c) are sub-models with only a subset of pathways.}
  \Description{A conceptual diagram showing the information pathways hypothesis.}
  \label{fig:hypothesis}
\end{figure}

\subsection{The Information Pathways Hypothesis}
The information pathways hypothesis is conceptually demonstrated in Fig.~\ref{fig:hypothesis}. We define a \emph{communication channel} $c_i$ as a series of self-attention based connections over multiple layers that let one element of the input gather information from another element. Each element may use many such channels to gather information from the context, i.e., other elements. A set of such connections (which may overlap) that can form a proper representation $e_i$ of a given element is called an \emph{information pathway} $P_i$. Multiple pathways may work together to form an embedding, but they can work independently as well, and can also be trained independently. The attention mechanism ensures that multiple sampled pathways are properly aggregated. If a pathway is sampled partially, it may introduce some noise in the aggregation. However, if the signals from the fully sampled pathways are strong enough, the network can ignore this noise (similar to a few weak models in an ensemble of mostly strong models). We define a \emph{sub-model} as one that uses only a subset of the pathways $P_i$ as in Fig.~\ref{fig:hypothesis} (b) and (c). A randomly sampled sub-model can be trained instead of the full model, which trains the sampled subset of the pathways. Even if a pathway is not sampled at a given step, it is trained indirectly because it shares weights with the sampled pathways. If a pathway positively contributes to the generalization performance of a transformer we call it an \emph{important information pathway}. With a proper sampling scheme, over multiple training steps, we can sample sub-models that cover most of the important information pathways. This is the key idea behind the proposed SSA method, which can efficiently sample the important pathways during training. Then, at inference time, we can use the full model to get the best performance, or we can use a set of sub-models to form an ensemble, which we call an \emph{attention self-ensemble}. This ensemble often produces more robust predictions than the full model, because of the regularization effect of the sampling process.

\subsection{Stochastically Subsampled Self-Attention} \label{sec:ssa}
In the self-attention process, all of the $N$ input elements form keys and values, and again all of the $N$ input elements form queries, which is responsible for the $N \times N$ shape of the self-attention matrix, and corresponding quadratic computational cost. To efficiently subsample the self-attention matrix we decouple the elements forming keys and values, which we call source elements, and the ones forming queries, which we call target elements. In our subsampling scheme, all of the elements in the input serve as targets, but each target only attends to a random subset of sources. That is, the queries $q_i$ are formed for all $i$ but each of them attends to key-value pairs $(k_j,v_j)$ for a random subset of $j$'s.
During sampling, the inclusion of a particular source multiple times for a given target is redundant. To avoid this, we ensure the sources are sampled without replacement for each target element. We propose two forms of SSA: i) {\em Unbiased SSA}, and ii) {\em Locally biased SSA}.

\begin{figure}[!t]
  \centering
  \includegraphics[width=\columnwidth]{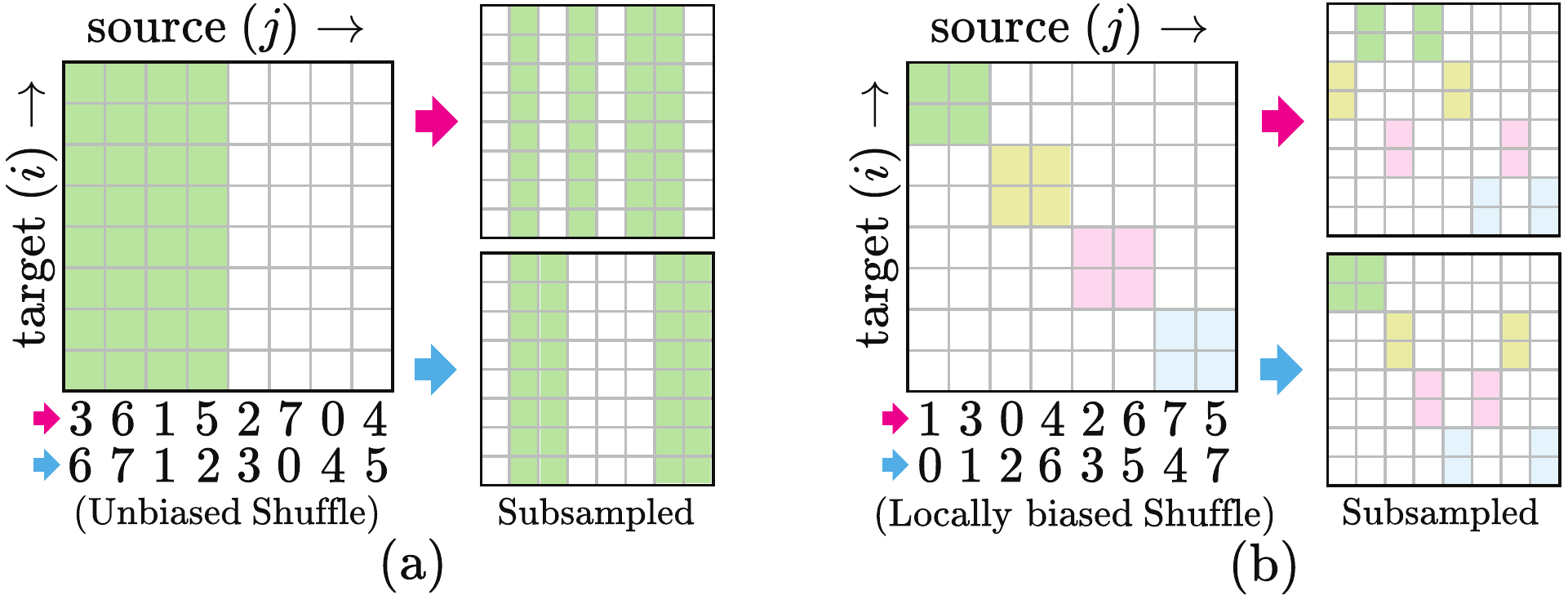}
  \caption{\small (a) Unbiased SSA uses unbiased source shuffling with truncation, (b) locally biased SSA uses locally biased source shuffling and windowed attention. Different attention patterns result from shuffling source indices (red and blue).}
  \Description{A conceptual demonstration of the two types of Stochastically Subsampled Self-Attention (SSA). }
  \label{fig:subsample}
\end{figure}

\smallskip\noindent{\bf Unbiased SSA:} In the first implementation of SSA shown in Algorithm~\ref{alg:unbiased}, we simply shuffle the sources in a random (unbiased) order (in line 1: randperm$(N)$), and truncate to keep only the first $k$ elements, as shown in Fig.~\ref{fig:subsample}(a). 
By subsampling $k$ sources for each target, unbiased SSA reduces the complexity of the self-attention process from $O(N^2)$ to $O(Nk)$.

\begin{algorithm}[!t]
  \caption{Unbiased SSA} \label{alg:unbiased}
  \begin{algorithmic}[1]
    \Require
    \small Subsampled length $k \in \mathbb{N}$;
    embeddings $\mathbf{X} \in \mathbb{R}^{N \times d}$;
    query projection matrix $\mathbf{W_q} \in \mathbb{R}^{d \times d_k}$;
    key projection matrix $\mathbf{W_k} \in \mathbb{R}^{d \times d_k}$;
    value projection matrix $\mathbf{W_v} \in \mathbb{R}^{d \times d_v}$;
    bias matrix $\mathbf{B} \in \mathbb{R}^{N \times N}$

    \Ensure
    \small Attention head $\mathbf{H} \in \mathbb{R}^{N \times d_v}$

    \State \small $\mathcal{P} \gets \mathrm{randperm}(N)$
    \Comment{\small Random permutation of indices: $N$}

    \State \small $\mathcal{\tilde P} \gets \mathcal{P}[0:k]$
    \Comment{\small Truncation: $k$}

    \State \small $\mathbf{X_{target}} \gets \mathbf{X}$
    \Comment{\small Target elements: $N \times d$}

    \State \small $\mathbf{X_{source}} \gets \mathbf{X}[\mathcal{\tilde P},:]$
    \Comment{\small Unbiased source sampling: $k \times d$}

    \State \small $\mathbf{Q} \gets \mathbf{X_{target}} \mathbf{W_q}$
    \Comment{\small Query projection: $N \times d_k$}

    \State \small $\mathbf{K} \gets \mathbf{X_{source}} \mathbf{W_k}$
    \Comment{\small Key projection: $k \times d_k$}

    \State \small $\mathbf{V} \gets \mathbf{X_{source}} \mathbf{W_v}$
    \Comment{\small Value projection: $k \times d_v$}

    \State \small $\mathbf{\tilde B} \gets \mathbf{B}[:,\mathcal{\tilde P}]$
    \Comment{\small Subsample bias matrix: $N \times k$}

    \State \small $\mathbf{H} \gets \mathrm{softmax}\left(\frac{\mathbf{Q} \mathbf{K}^T}{\sqrt{d_k}} + \mathbf{\tilde B}\right)\mathbf{V}$
    \Comment{\small Self-attention: $N \times d_v$}
    \\
    \Return $\mathbf{H}$
\end{algorithmic}
\end{algorithm}

\smallskip\noindent{\bf Locally Biased SSA:}
Here, we form local windows for both sources and targets, as shown in Algorithm~\ref{alg:biased}. If both the source and target windows contain local patches of elements, then attention is confined within that window. However, if we rearrange the sources in a locally biased random order (in line 1: localrandperm$(N, w, \sigma)$), then the targets can attend to elements beyond their own window, possibly from the entire input with a non-zero probability (Fig.~\ref{fig:subsample}(b)). 
By subsampling $w$ local windows, locally biased subsampling pairs each target with only $N/w$ sources, reducing the complexity from $O(N^2)$ to $O(N^2/w)$. 

\begin{algorithm}[!t]
  \caption{Locally Biased SSA} \label{alg:biased}
  \begin{algorithmic}[1]
    \Require
    \small Number of local windows $w \in \mathbb{N}$;
    Standard deviation of local bias $\sigma \in \mathbb{R}$;
    embeddings $\mathbf{X} \in \mathbb{R}^{N \times d}$;
    query projection matrix $\mathbf{W_q} \in \mathbb{R}^{d \times d_k}$;
    key projection matrix $\mathbf{W_k} \in \mathbb{R}^{d \times d_k}$;
    value projection matrix $\mathbf{W_v} \in \mathbb{R}^{d \times d_v}$;
    bias matrix $\mathbf{B} \in \mathbb{R}^{N \times N}$

    \Ensure
    \small Attention head $\mathbf{H} \in \mathbb{R}^{N \times d_v}$

    \State \small $\mathcal{P} \gets \mathrm{localrandperm}(N, w, \sigma)$
    \Comment{\small Locally biased random}
    \state \small \ \Comment{\small permutation of indices: $N$}

    \State \small $\mathbf{X_{target}} \gets \mathbf{X}$
    \Comment{\small Target elements: $N \times d$}

    \State \small $\mathbf{X_{source}} \gets \mathbf{X}[\mathcal{P},:]$
    \Comment{\small Reindex source elements: $N \times d$}

    \State \small $\mathbf{Q} \gets \mathbf{X_{target}} \mathbf{W_q}$
    \Comment{\small Query projection: $N \times d_k$}

    \State \small $\mathbf{K} \gets \mathbf{X_{source}} \mathbf{W_k}$
    \Comment{\small Key projection: $N \times d_k$}

    \State \small $\mathbf{V} \gets \mathbf{X_{source}} \mathbf{W_v}$
    \Comment{\small Value projection: $N \times d_v$}

    \State \small $\mathbf{\tilde B} \gets \mathbf{B}[:,\mathcal{P}]$
    \Comment{\small Reindex bias matrix: $N \times N$}

    \State \small $\mathbf{Q_w} \gets \mathrm{window}\left(\mathbf{Q}, w\right)$
    \Comment{\small Window query: $w \times \sfrac{N}{w} \times d_k$}

    \State \small $\mathbf{K_w} \gets \mathrm{window}\left(\mathbf{K}, w\right)$
    \Comment{\small Window key: $w \times \sfrac{N}{w} \times d_k$}

    \State \small $\mathbf{V_w} \gets \mathrm{window}\left(\mathbf{V}, w\right)$
    \Comment{\small Window value: $w \times \sfrac{N}{w} \times d_v$}

    \State \small $\mathbf{\tilde B_w} \gets \mathrm{window_{diag}}\left(\mathbf{\tilde B}, w\right)$
    \Comment{\small Window bias along diagonal}
    \state \small \ \Comment{\small blocks: $w \times \sfrac{N}{w} \times \sfrac{N}{w}$}

    \State \small $\mathbf{H_w} \gets \mathrm{softmax}\left(\frac{\mathbf{Q_w} \mathbf{K_w^T}}{\sqrt{d_k}} + \mathbf{\tilde B_w}\right)\mathbf{V_w}$
    \Comment{\small Self-attention: $w \times \sfrac{N}{w} \times d_v$}

    \State \small $\mathbf{H} \gets \mathrm{flattenwindow}(\mathbf{H_w})$
    \Comment{\small Flatten windowed attention}
    \state \small \ \Comment{\small head: $N \times d_v$}
    \\
    \Return $\mathbf{H}$
\end{algorithmic}
\end{algorithm}

Unbiased SSA is very easy to implement, but in our experiments, we found that locally biased SSA works better both in terms of model performance and efficiency.
We pair the same set of sources with all targets for unbiased SSA, or within each window for locally biased SSA. This ensured that we can use highly optimized dense tensor multiplications for attention. Also, we use the same set of sources for all attention heads within a layer. This allows us to perform SSA by simply reindexing the embeddings and the bias matrix, followed by an unaltered/windowed multihead attention. We also use the same reindexing within each mini-batch, although, in a distributed data-parallel setting, each worker may have different indices. 
Both SSA algorithms can be implemented in any modern deep-learning framework in a few lines of code, without use of sparse tensor operations or custom GPU kernels. 

\begin{figure}[!t]
  \centering
  \includegraphics[width=0.8\columnwidth]{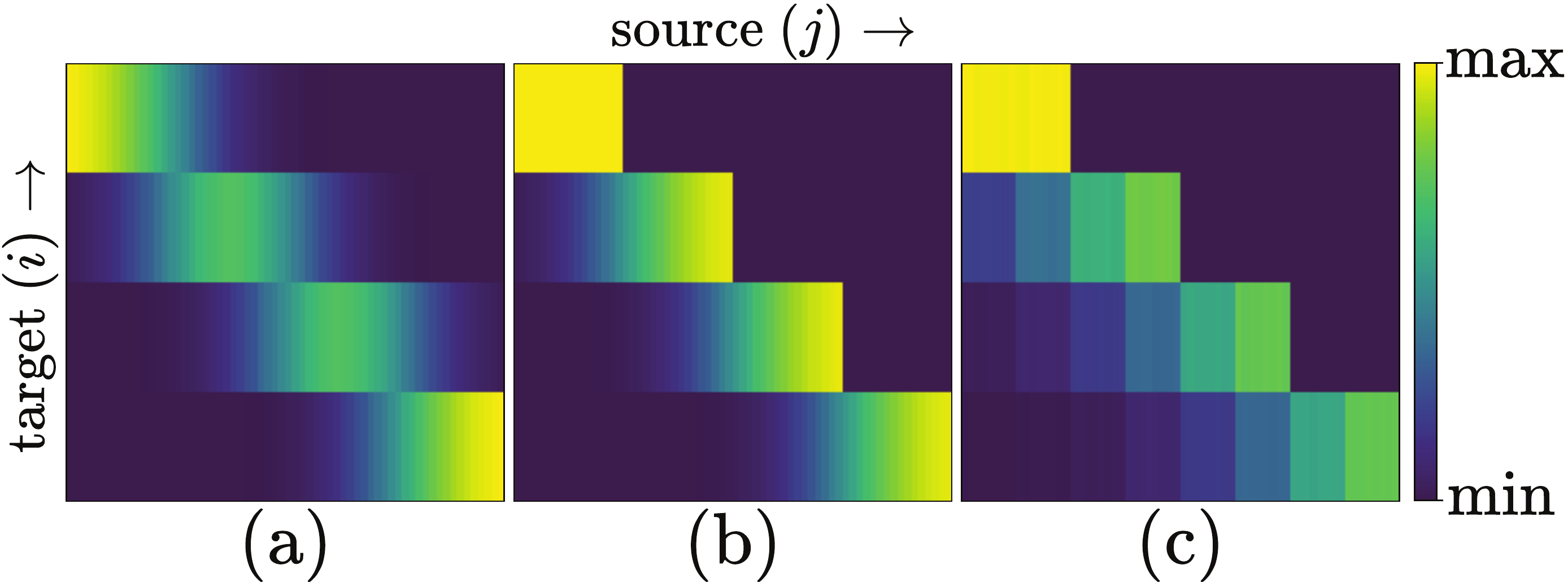}
  \caption{\small Sampling probability of the self-attention matrix for different types of locally biased sampling: (a) gaussian, (b) causal gaussian, and (c) causal gaussian for 2D grids with vertical windows.}
  \Description{A diagram showing the sampling probability of the self-attention matrix for different types of locally biased sampling by the SSA algorithm.}
  \label{fig:distributions}
  \vspace{5\lineskip}
  \includegraphics[width=\columnwidth]{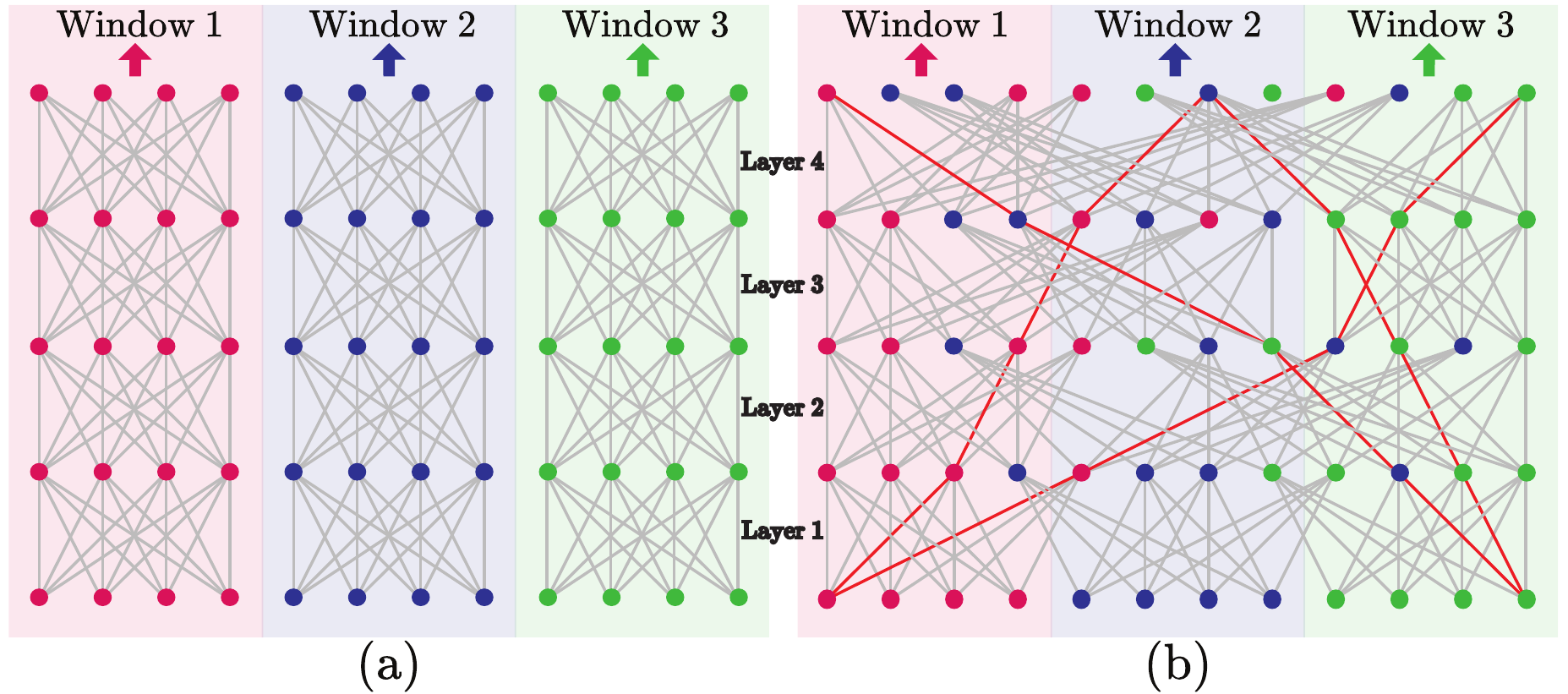}
  \caption{\small Windowed attention with (a) no source shuffling vs.\ (b) locally biased source shuffling -- some sources move to other windows forming long-range connections, some of which are shown in red.}
  \Description{A conceptual diagram showing the implications of locally biased SSA on the subsampled connectivity patterns in a deep network.}
  \label{fig:windows}
  \vspace{-0.5em}
\end{figure}

We implement locally biased shuffling (localrandperm $(N, w, \sigma)$) by generating permutation indices whereby each index can shift around its original position with a gaussian probability distribution. We do this by adding gaussian noise to the indices:
\begin{align}
  \mathcal{P} = \mathrm{argsort}\left(\{1+n_1, 2+n_2, 3+n_3, \ldots, N+n_N\}\right)  \label{eq:locperm}
\end{align}
where $n_i \sim \mathcal{N}(0, \sigma^2)$, and
the standard deviation $\sigma$ controls the amount of local bias. A lower value of $\sigma$ results in more local bias, whereas $\sigma \to \infty$ would lead to no local bias. The resultant subsampling distribution is shown in Fig.~\ref{fig:distributions} (a), where we can see that the sampling probabilities are concentrated towards the diagonal of the self-attention matrix. For generative tasks, we use a causal version of locally biased SSA, where the permutation indices are resampled for each window, and are constrained to be from 0 to the end of the window. The resulting sampling distribution is shown in Fig.~\ref{fig:distributions} (b). For 2D grids, such as images we perform shuffling both horizontally and vertically. For image generation, we partition the grid vertically into $w$ windows. The resultant distribution after locally biased shuffling and windowing is shown in Fig.~\ref{fig:distributions} (c). Here we have flattened the grid row-by-row.

In Fig.~\ref{fig:windows} we show the implications of locally biased SSA on the subsampled connectivity patterns in a deep network. Simply performing windowed attention in each layer would isolate each window as in Fig.~\ref{fig:windows} (a). This is why local self-attention methods use either overlapping \citep{beltagy2020longformer,child2019generating} or shifted windows \citep{liu2021swin} to ensure connectivity across windows. Instead, we rely on the stochasticity of the sampling process for inter-window connectivity. We can see in Fig.~\ref{fig:windows} (b) that with locally biased SSA, after a few layers, we have long-distance connections across window boundaries with a non-zero probability while maintaining the same level of sparsity. Note that methods like BigBird \citep{zaheer2020big} achieve this by a combination of local and random attention, which is kept \emph{fixed} during training and inference. In contrast, the sparsity patterns in SSA are {\em resampled at every training step}, and we can fall back to dense attention during inference. Also, slowly reducing local bias (increasing $\sigma$) in the deeper layers leads to better generalization. We hypothesize that within the important information pathways, local connections are formed predominantly in the shallower layers while long-range connections are formed in deeper layers. For a given sparsity budget, locally biased SSA can sample these pathways with a higher probability than unbiased SSA. This is why locally biased SSA can achieve better performance at a lower training cost.

\subsection{Fine-tuning and Inference}
After training with SSA, we fall back to dense attention at inference time, which ensures that the network leverages all information pathways to produce the best output. This is analogous to the rescaling/renormalization in dropout \citep{srivastava2014dropout} at inference time. In our case, the attention process ensures that the contributions of the pathways are properly aggregated via its weighted averaging process, so that no manual rescaling is required. We call this \textbf{attention-based renormalization}. Often, no extra training is required to ensure proper renormalization and good performance at inference time. However, especially when we apply a high sparsity during training, the network may need some extra adjustment to ensure proper renormalization. A small amount of fine-tuning with dense attention at the end of training is sufficient to ensure good performance at inference time. This is done in the last few epochs ($\le 10\%$ of the total epochs). This method falls within the category of curriculum learning \citep{bengio2009curriculum} strategies such as \citep{press2020shortformer,li2021curriculum}. Although training can be significantly slower without SSA, since this is done only for a few epochs, the overall training time is not significantly affected. This fine-tuning step is not required when we use only moderately sparse attention ($\le 50\%$ sparsity) during training, because the network does not face a drastic distribution shift from the training to the inference time in this case.

\subsection{SSA-based Attention Self-Ensembling}
Generation of an ensemble of sub-models using SSA is as simple as performing SSA at inference time on the trained model, drawing multiple sample outputs for the same input and taking an aggregation of the predictions. Although this method leverages the same model weights for each sample prediction, SSA draws a {\em random subsampling pattern} each time, producing a set of sub-models that only vary in their attention patterns. We use an average of the predicted probabilities of the sub-models for generative or classification tasks, or a mean of the predicted values for regression tasks. Surprisingly, we found that the average predictions of these sub-models can be more robust and generalizable than that of the full model if SSA is performed meticulously (i.e., if the SSA hyperparameters are chosen carefully). This shows that the full model may suffer from over-capacity, and thus overfit the training data. Even at inference time, SSA can uncover lower capacity models which may have more generalizable traits such as prioritizing long-distance dependencies over short-distance ones. Although SSA-based self-ensembling works best when the model is trained with SSA, we found that it can work with a model trained with vanilla dense attention as well, often matching or even outperforming the dense model. Also, the fact that an ensemble of sub-models can be as performant as the full model shows that the transformer can be thought of as an ensemble of these sub-models with the attention mechanism aggregating/merging them into a single model. {\em This also gives evidence in favor of the information pathways hypothesis}, by showing that sub-models can be formed from a subset of the connectivities, indicating the existence of alternative information pathways in the transformer which can operate independently.

SSA-based attention self-ensembling works best with SSA training, and can often serve as an alternative to fine-tuning or dense-attention fallback. In this case SSA is performed both during training and inference. As a result, we have the same distribution of subsampled attention, so the network does not need to readjust to a different distribution at inference time. Also, the SSA inference for each sub-model can be much less costly and less memory intensive than the full model which uses dense attention. Although we need to draw multiple samples, this process is embarrassingly parallel and can be easily done on separate workers (CPUs/GPUs/nodes) followed by an aggregation step. All sub-models in a self-ensemble share the same set of parameters, so the total number of parameters is the same as that of the full model. There is no added training cost since we train a single model with SSA. This makes it easier to train and deploy the ensemble. As such, attention self-ensemble is a more general concept and can potentially be used with other forms of stochastic subsampling methods (e.g., attention dropout), and also for uncertainty estimation, similar to \citep{gal2016dropout}.

\section{Experiments}
We explore the effectiveness of SSA for various tasks involving transformers. We experiment with different types of data and both generative and discriminative tasks, such as generative modeling of text, image generation, image classification and graph regression. Our experiments cover different granularities of input data as well, e.g., for text, we consider both word-level and character-level inputs, for images we consider both pixel-level and patch-level inputs and for graphs we process individual node-level inputs. Also, we explore different scales such as relatively smaller-scale CIFAR-10 \citep{krizhevsky2009learning} image dataset, medium-scale Enwik8 \citep{mahoney2011large} and WikiText-103 \citep{merity2016pointer} text datasets and large scale ImageNet-1K \citep{deng2009imagenet} and PCQM4Mv2 \citep{hu2021ogb} molecular graph datasets. We used the PyTorch \citep{paszke2019pytorch} library for our experiments. The training was done in a distributed manner with mixed-precision computation on up to 4 nodes, each with 8 NVIDIA Tesla V100 GPUs (32GB RAM/GPU), and two 20-core 2.5GHz Intel Xeon CPUs (768GB RAM). More details about the hyperparameters and the training procedure are provided in the Appendix. 
Our code is available at \url{https://github.com/shamim-hussain/ssa}.

\subsection{Generative Language Modeling}
Our language modeling experiments showcase the application of SSA to generative modeling of text data, and its ability to handle long-range dependencies. We experiment on the WikiText-103 and the Enwik8 datasets. The WikiText-103 \citep{merity2016pointer} dataset contains a diverse collection of English Wikipedia articles with a total of 103 million word-level tokens. This dataset has been extensively used as a long-range language modeling benchmark. The Enwik8 \citep{mahoney2011large} dataset contains the first 100 million bytes of unprocessed Wikipedia text. This dataset has been used as a benchmark for byte-level text compression. For both these datasets, we used the 16-layer transformer decoder of \citet{press2021train} which uses ALiBi relative positional encodings. We used an input length of 3072 tokens for WikiText-103. We made minor changes to the architecture and training procedure (refer to the Appendix), which allow us to train the model much faster on 32 V100 GPUs, within 9 hours, compared to the 48 hours required by \citet{press2021train}, while still yielding comparable perplexity. We achieve validation and test perplexities of 17.14 and 17.98, with a sliding window inference (overlap length 2048), compared to 16.96 and 17.68 of \citet{press2021train} with vanilla dense attention training. We call this S0 (since SSA was used in 0 layers) and use this as a baseline for SSA results. On Enwik8, we get validation and test BPB (bits per byte) of 1.052 and 1.028 with a sliding window inference (overlap length 3072), which we use as the baseline (i.e., S0). We could not find a comparable dense attention implementation; \citet{al2019character} achieve a test BPB of 1.06 with a very deep 64-layer transformer. A local transformer achieves a test BPB of 1.10, whereas specialized architectures such as \citep{dai2019transformer,roy2021efficient} use a longer input length to achieve a test BPB of 0.99, which is comparable to ours. We could train only up to an input length of 4096 with dense attention without gradient accumulation/checkpointing, so we experiment with this input length.

Our experiments are designed to show the effectiveness of SSA in reducing training costs and also as a regularization method. We measure training cost in terms of Compute (FLOPs), Memory (GB) and Speed (steps/sec). We normalize these with respect to S0, to better represent comparative gains achieved with SSA (refer to the Appendix for unnormalized values). We primarily show results for locally biased SSA since it produces the best results, and leave the results for unbiased SSA as an ablation study (refer to the Appendix). We use the causal gaussian sampling scheme described in section \ref{sec:ssa}. We tune the SSA parameters $\sigma$ (in Eq. \ref{eq:locperm}) in different layers for the best validation set results. We applied different numbers of windows with locally biased SSA to achieve different levels of sparsity and regularization, both of which increase with the number of windows. For example, with 4 windows we reduce the attention cost 4 times by only sampling 25\% of the self-attention matrix. This is denoted with a suffix `-L4' (Locally biased with 4 windows). We mainly apply SSA to all 16 transformer layers (S16), but we found that sometimes better results can be achieved by leaving the first few layers unsampled, at the cost of some efficiency. For example, we use S12 to denote that SSA has been applied only to the last 12 layers. Also, we produced results for the Fine-Tuning (+FT) scheme where we turn off SSA in the last 10\% of the training epochs and fine-tune the model for dense attention, which leads to better results. For additional fine-tuning, we report the total compute, but average speedup and memory consumption over the training epochs.

\begin{table}[!t]
  \centering
  \caption{\small Results on language modeling tasks on WikiText-103 and Enwik8. \textbs{Red:} best model, \textgd{Violet:} good model; C/M/S: normalized Compute/Memory/Speedup; Ppl.: perplexity; BPB: bits per byte; arrow indicates if higher or lower is better.}
  \label{tab:text_res}
  \scalebox{0.74}[.76]{
    \newcommand{\cmscol}{\textbf{C}$\downarrow$ / \textbf{M}$\downarrow$ / \textbf{S}$\uparrow$}
\begin{tabular}{l|cc|cc}
    \toprule
                          &\multicolumn{2}{c|}{\textbf{WikiText-103 (Gen.)}}                        &\multicolumn{2}{c}{\textbf{Enwik8 (Gen.)}}                                         \\
                          &\multicolumn{2}{c|}{(\#Layers=16, \#Params=247M)}                        &\multicolumn{2}{c}{(\#Layers=16, \#Params=202M)}                                   \\
    \textbf{Model*}       & \textbf{dev/test Ppl.}$\downarrow$   & \cmscol                          & \textbf{dev/test BPB}$\downarrow$    & \cmscol                                    \\ \midrule                                                                                       
    S0({\small Dense})    &  17.14 / 17.98                       &          1.00 / 1.00 / 1.00      &  1.052 / 1.028                       &          1.00 / 1.00 / 1.00                \\ \midrule                                                                                                               
    S16-L2                &  17.12 / 17.84                       &          0.83 / 0.74 / 1.15      &  1.052 / 1.028                       &          0.80 / 0.67 / 1.34                \\ 
    +FT                   &  \textgd{16.95} / \textgd{17.68}     &          0.85 / 0.77 / 1.13      &  \textgd{1.050} / \textgd{1.026}     &          0.82 / 0.71 / 1.30                \\ \midrule                                                                                                               
    S16-L4                &  17.39 / 18.13                       &          0.75 / 0.62 / 1.31      &  1.081 / 1.058                       &  \textbs{0.70 / 0.51 / 1.64}               \\ 
    +FT                   &  \textbs{16.91} / \textbs{17.60}     &          0.78 / 0.65 / 1.27      &  1.052 / 1.029                       &  \textgd{0.73 / 0.56 / 1.55}               \\ \midrule                                                                                                               
    S12-L4                &  17.29 / 17.95                       &          0.81 / 0.71 / 1.22      &  \textgd{1.047} / \textbs{1.024}     &          0.78 / 0.64 / 1.48                \\ 
    +FT                   &  17.09 / 17.86                       &          0.83 / 0.74 / 1.20      &  \textbs{1.044} / \textbs{1.024}     &          0.80 / 0.67 / 1.41                \\ \midrule                                                                                                              
    S16-L6                &  17.49 / 18.30                       &  \textgd{0.72 / 0.57 / 1.39}     &  \multicolumn{2}{l}{\small \textbf{*S<$\ell$>-L<$w$>:} Locally biased SSA}        \\ 
    +FT                   &  17.09 / 17.86                       &          0.75 / 0.62 / 1.34      &  \multicolumn{2}{l}{\small on the last $\ell$ layers with $w$ windows}            \\ \cmidrule{1-3}                                                                                                              
    S16-L8                &  17.94 / 18.69                       &  \textbs{0.71 / 0.55 / 1.42}     &  \multicolumn{2}{l}{\small \textbf{+FT:} Finetuned without SSA for the}           \\
    +FT                   &  17.20 / 17.92                       &          0.74 / 0.60 / 1.36      &  \multicolumn{2}{l}{\small last 10\% epochs}\\
    \bottomrule                                                                                            
\end{tabular}
    }
\end{table}

The results are presented in Table \ref{tab:text_res}. On WikiText-103 we achieve the best result with S16-L4 after fine-tuning. Here, SSA is used in all layers with 4 windows, which corresponds to only 25\% of attention being sampled during the majority of the training. We achieve a significant improvement over the baseline (S0) due to the regularization effect of SSA, while also achieving a 1.27x speedup in training, 22\% reduction in compute and 35\% reduction in memory cost. This method also achieves competitive results on Enwik8, but the best result is achieved by S12-L4, where we leave the first 4 layers unsampled. We think this is due to the higher granularity of character-level data, which makes the locally biased SSA algorithm less effective in predicting the attention patterns in the shallower layers. S12-L4 achieves the best result even without fine-tuning and also has 1.48x speedup in training, 22\% reduction in compute and 36\% reduction in memory cost. Both S16-L2 and S12-L4 achieve good results even without fine-tuning, which shows that the requirement for fine-tuning arises mainly due to highly sparse sampling. We can reduce the training cost further by using sparser subsampling, for example, with S16-L6 or S16-L8 but this comes at the cost of slightly worse results, even after fine-tuning. We believe this is because, at very high sparsity levels, some of the important pathways remain undertrained, which is corrected only slightly by fine-tuning. Also, at this point, other parts of the network become the bottleneck rather than self-attention, which leads to diminishing returns in terms of training cost reduction.

\begin{figure}[!t]
  \centering
  \includegraphics[width=\columnwidth]{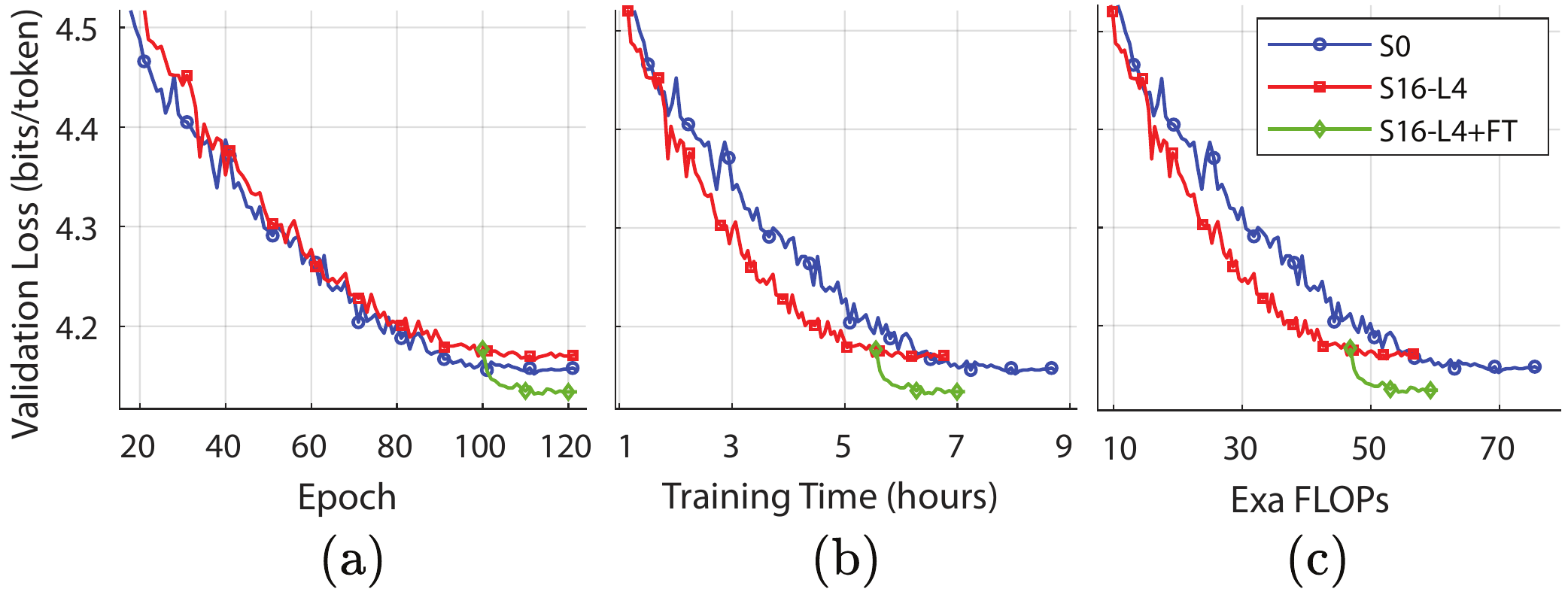}
  \caption{\small Validation loss vs training (a) epochs, (b) time and (c) compute for the WikiText-103 experiment, with (\textcolor{red}{red}) and without (\textcolor{blue}{blue}) SSA and with fine-tuning (\textcolor{green}{green}) which begins at epoch 100.}
  \Description{A graph showing the validation loss vs training epochs, time, and compute with and without SSA and with fine-tuning.}
  \label{fig:training}
\end{figure}

In Fig.~\ref{fig:training} we see how training with SSA progresses compared to dense attention training. From Fig.~\ref{fig:training}(a) we see that the validation loss of S16-L4 closely follows that of S0 for most of the training in terms of the number of steps. This verifies our claim that the information pathways can be trained independently by showing that even when we are sampling a small subset (25\%) of the pathways, training progresses naturally. However, in terms of both wall time and compute, the validation loss of S16-L4 drops much faster than S0. The validation loss plateaus at a slightly higher value than that of S0, but with a slight fine-tuning in the end, it falls even below that of S0. Also, even with fine-tuning, training finishes significantly earlier than S0. Thus, compared to dense attention (S0), SSA delivers significant improvements in performance and efficiency.

\subsection{Image Generation and Classification}
While some previous works only focus on reducing the cost of training only for generative language modeling \citep{press2020shortformer,li2021curriculum}, we show the generality of our method by also applying it to image generation and classification tasks. We target the unconditional sub-pixel level image generation task on CIFAR-10 \citep{krizhevsky2009learning}, which contains 60,000 tiny 32x32x3 images from 10 classes. Each image is flattened into a sequence of length 3072 and fed to a transformer decoder, which serves as an autoregressive model. We get a validation BPD (bits per dimension) of 2.789 with dense attention training which we denote as the baseline S0. We could not find a comparable dense attention result in the literature, but some specialized architectures such as \citep{child2019generating,roy2021efficient} have reported comparable results. Our results are presented in Table \ref{tab:img_res} (left). We see that with fine-tuning, SSA achieves a slightly better result than dense training (S0) while achieving 1.22x speedup, saving 23\% compute and 42\% memory. Without fine-tuning, it achieves a slightly worse result but almost halves the memory required for training, which is particularly beneficial for high-resolution image generation.

\begin{table}[!t]
  \centering
  \caption{\small Image generation results on CIFAR-10 and image classification results on ImageNet-1K. BPD: bits per dimension; Acc.: top-1 accuracy. \textbs{Red:} best model, \textgd{Violet:} good model.}
  \label{tab:img_res}
  \scalebox{0.74}[.76]{
    \newcommand{\cmscol}{\textbf{C}$\downarrow$ / \textbf{M}$\downarrow$ / \textbf{S}$\uparrow$}
\begin{tabular}{lcc|lcc}
    \toprule
    \multicolumn{3}{c|}{\textbf{CIFAR-10 (Gen.)}}                                              & \multicolumn{3}{c}{\textbf{ImageNet-1K (Class.)}}                                              \\ 
    \multicolumn{3}{c|}{(\#Layers=16, \#Params=203M)}                                          & \multicolumn{3}{c}{(Swin-T, \#Layers=12, \#Params=28M)}                                        \\               
      \textbf{Model}      & \textbf{BPD}$\downarrow$          & \cmscol                        &   \textbf{Model*}        & \textbf{Acc.}$\uparrow$           & \cmscol                         \\ \midrule
      S0({\small Dense})  &  \textgd{2.789}                   &          1.00 / 1.00 / 1.00    &   W7-S0({\small Dense})  &          81.19\%                  &  \textbs{0.90 / 0.70 / 1.14}    \\ \midrule
      S16-L4              &  2.796                            &  \textbs{0.75 / 0.53 / 1.25}   &   W14-S10-L4             &          80.56\%                  &  \textgd{0.90 / 0.73 / 1.13}    \\               
      +FT                 &  \textbs{2.774}                   &  \textgd{0.77 / 0.58 / 1.22}   &   +FT                    &          81.15\%                  &          0.91 / 0.76 / 1.13     \\ \midrule               
    \multicolumn{3}{l|}{}                                                                      &   W14-S6-L4              &          81.23\%                  &          0.97 / 0.91 / 1.05     \\                
    \multicolumn{3}{l|}{}                                                                      &   +FT                    &  \textgd{81.60\%}                 &          0.97 / 0.92 / 1.05     \\ \cmidrule{4-6}               
    \multicolumn{3}{l|}{\small \textbf{*W<$\omega$>-\dots:} Window-size of Swin-T = $\omega$}  &   W14-S0({\small Dense}) &  \textbs{81.89\%}                 &          1.00 / 1.00 / 1.00     \\                                                                                                                                                                                                                                                                                                                                                                                                                                                                                                          
    \bottomrule                                                                                            
\end{tabular}

    }
\end{table}

Beyond generative tasks, we also explore the usefulness of SSA for discriminative tasks such as the large-scale image classification task on the ImageNet-1K dataset \citep{deng2009imagenet} which contains 1.28 million images from 1000 classes. It is customary to train transformers on image patches for classification. Instead of the vanilla Vision Transformer \citep{dosovitskiy2020image}, we use the Swin Transformer \citep{liu2021swin} because it achieves state-of-the-art results on ImageNet-1K when trained from scratch. Additionally, we aim to investigate SSA's applicability to locally dense attention based architectures such as the Swin Transformer, which uses a shifted window based local attention mechanism enabling efficient handling of smaller patches (e.g., 4x4). We use the Swin-Tiny model with 12 layers and 28 million parameters and an input resolution of 224x224 in our experiments, and report the top-1 accuracy on the validation set. To demonstrate the usefulness of SSA, we use window sizes of 7x7 and 14x14, denoted by W7 and W14 respectively. A larger window uses more attention and achieves better results, but also requires more compute and memory. The results are presented in Table \ref{tab:img_res} (right). To apply SSA we subdivide each window into 4 sub-windows (L4). With SSA applied to 10 layers (the last two layers have a resolution of 7x7, where further sub-division is not possible), we can train a W14 model with almost the same training cost as a W7 model. However, even with fine-tuning, we cannot achieve better results than W7. Only by excluding the first 4 layers from SSA and fine-tuning, we attain better accuracy than W7. This accuracy is, however, slightly less than that of W14-S0, but we achieve this at a lower training cost. We believe that this is because the shifted window based attention mechanism is inherently more local than global attention, limiting the regularization effect of locally biased SSA. Moreover, attention is no longer the primary bottleneck. Hence, the savings due to SSA are only incremental. However, SSA can still be utilized to trade off accuracy for training cost, as evidenced by the 3\% compute and 8\% memory savings, as well as the 5\% speedup over the locally dense model.

\subsection{Molecular Graph Regression}
We further show the generality of our method by applying SSA to molecular graph data on the PCQM4Mv2 quantum chemical dataset \citep{hu2021ogb}. Also, we wanted to demonstrate its applicability to newly proposed Graph Transformers \citep{ying2021transformers,hussain2022global,park2022grpe}, which use global self-attention. The PCQM4Mv2 dataset contains 3.8 million molecular graphs, and the target task is to predict a continuous valued property, the HOMO-LUMO gap, for each molecule. For this task, we use the Edge-augmented Graph Transformer (EGT) \citep{hussain2022global}. We experiment with an ablated variant of EGT called EGT-Simple since it approximately achieves the same performance on PCQM4Mv2 while also being simpler to apply SSA to, but for brevity, we will call this model EGT. We experiment on the EGT\textsubscript{small} model with 11 million parameters and 6 layers, and report the mean absolute error (MAE) on the validation set. We achieve a baseline MAE of 0.0905, as reported in \citep{hussain2022global} without SSA, which we call S0.

\begin{table}[!t]
  \centering
  \caption{\small Graph regression results on PCQM4Mv2 dataset. MAE: mean absolute error.}
  \label{tab:graph_res}
  \scalebox{0.77}{
    \newcommand{\cmscol}{\textbf{C}$\downarrow$ / \textbf{M}$\downarrow$ / \textbf{S}$\uparrow$}
\begin{tabular}{l|cc|cc|l}
    \toprule
      \multicolumn{1}{l}{}   & \multicolumn{5}{c}{\textbf{PCQM4Mv2 (Regr.)}}                                                                                                                                            \\
      \multicolumn{1}{l}{}   & \multicolumn{5}{c}{(EGT, \#Layers=6, \#Params=11M)}                                                                                                                                      \\
      \multicolumn{1}{l|}{}  & \multicolumn{2}{c|}{\textbf{w/o FT}}                        & \multicolumn{2}{c}{\textbf{+ FT}}                                                &                                         \\
      \textbf{Model*}        & \textbf{dev MAE}$\downarrow$  &\textbf{Compute}$\downarrow$ & \textbf{dev MAE}$\downarrow$  & \multicolumn{1}{c}{\textbf{Compute}$\downarrow$} &                                         \\ \midrule
      S0({\small Dense})     &          0.0905               &          1.00               &                               &  \multicolumn{1}{c}{}                            &                                         \\ \midrule
      S6-U10                 &          0.0907               &          0.96               &              0.0895           &          0.97                                    &  \small \textbf{*S<$\ell$>-U<$x$>:}     \\ 
      S6-U20                 &          0.0895               &          0.93               &      \textbs{0.0876}          &          0.94                                    &  \small Unbiased SSA                    \\ 
      S6-U30                 &          0.0904               &          0.89               &      \textbs{0.0876}          &          0.90                                    &  \small on the last $\ell$              \\ 
      S6-U40                 &          0.0930               &          0.86               &      \textgd{0.0879}          &          0.87                                    &  \small layers with                     \\ 
      S6-U50                 &          0.0964               &  \textbs{0.82}              &              0.0908           &  \textgd{0.84}                                   &  \small $x\%$ drop                      \\ 
    \bottomrule                                                             
\end{tabular}
    }
\end{table}
Graphs are fundamentally different from images and text due to their arbitrary topology and do not have a single simplistic notion of locality. To apply locally biased SSA we must partition the graph into equally sized local windows. There are different possible ways of doing it which may also involve the edge features. Further, we need to do locally biased source shuffling on graph nodes. Since this would require substantial further research, we instead show results for unbiased SSA on graphs, which is straightforward to implement as it does not rely on the notion of locality. We apply SSA to all layers (S6) and drop 10\%-50\% of source nodes randomly during training. For example, we use the suffix `-U20' to denote that 20\% of the source nodes are randomly dropped and we sample the remaining 80\%. We also report the result after fine-tuning without SSA for the last 10\% of the training epochs (+FT). The results are shown in Table \ref{tab:graph_res}. We see that the best results (MAE of 0.0876) are achieved for S6-U20 and S6-U30 with fine-tuning which is not only significantly better than the baseline (S0) but also requires around 10\% less compute (FLOPs). For this training, we could not tabulate the memory savings and speedup because in our implementation the data-loading of graphs becomes the bottleneck. We believe that the better results achieved by SSA on graphs are due to its regularization effect, which encourages the network to consider long-range interactions. However, unlike locally biased SSA, unbiased SSA cannot employ highly sparse attention without incurring a performance penalty, as evident from the results of S6-U50. At 50\% sparsity, the important pathways are rarely sampled and remain undertrained. We leave it as a future research direction to explore the use of locally biased SSA on graphs, which we believe will further improve the performance and efficiency of training.

\begin{table}[!t]
  \centering
  \caption{\small Self-ensembling results by locally biased SSA with 4 windows on WikiText-103 and Enwik8, produced with 50 samples for each input segment. Renormalized results are from Table \ref{tab:text_res}.}
  \label{tab:text_ens}
  \scalebox{0.77}{
    \begin{tabular}{l|cc|cc}
    \toprule
                       &\multicolumn{2}{c|}{\textbf{Wikitext-103}}                                              & \multicolumn{2}{c}{\textbf{Enwik8}}                                                           \\
                       &\multicolumn{2}{c|}{\textbf{dev/test Ppl.} $\downarrow$}                                & \multicolumn{2}{c}{\textbf{dev/test BPB} $\downarrow$}                                        \\
    \textbf{Model}     &\textbf{Renorm.}                                      &\textbf{Ensemble}                & \textbf{Renorm.}                                      &\textbf{Ensemble}                      \\ \midrule                                                                                                      
    S0({\small Dense}) &          17.14 / 17.98                               & 16.86 / 17.46                   &           1.052 / 1.028                               & 1.066 / 1.042                         \\ \midrule                                                                                                              
    S16-L4             &          17.39 / 18.13                               & \textgd{16.75} / \textgd{17.42} &           1.081 / 1.058                               & 1.086 / 1.062                         \\
    +FT                &          16.91 / 17.60                               & \textbs{16.54} / \textbs{17.18} &           1.052 / 1.029                               & 1.058 / 1.035                         \\\midrule                                                                                                              
    S12-L4             &          17.29 / 17.95                               & 16.89 / 17.60                   &   \textgd{1.047} / \textbs{1.024}                     & 1.050 / 1.029                         \\
    +FT                &          17.09 / 17.86                               & 16.80 / 17.51                   &   \textbs{1.044} / \textbs{1.024}                     & 1.055 / 1.033                         \\
    \bottomrule                                                                                                           
\end{tabular}%
    }
\end{table}

\subsection{Self-ensembling Results}
Once a transformer has been trained we can apply SSA at inference time, draw multiple sample predictions from the same input and aggregate them. This way the prediction is made by an ensemble of sub-models, sampled by SSA, which we call self-ensembling. The results of an average of 50 prediction samples drawn by locally biased SSA with 4 windows, which samples 25\% attention at each prediction instance for language modeling tasks, are shown in Table \ref{tab:text_ens}, and they are compared against their full-model counterpart, which we call renormalized results (since the network merges and normalizes the sub-models into a single model). For WikiText-103, we see that the self-ensembling results are significantly better than their renormalized counterparts. This is true even for S0, which was not trained with SSA but with vanilla dense attention. This shows that SSA-based self-ensembling can improve the performance of the model even when it is not trained with SSA. This also shows the existence of sub-models within a dense transformer, trained with dense attention, which is an implication of the information pathway hypothesis. Results are better when the model is trained with SSA and fine-tuning further improves the results. We think the better results are due to the higher generalizability of the constituent sub-models which take advantage of the local inductive bias and higher sparsity regularization. For Enwik8, however, the results are close to but not better than the renormalized counterparts. We think this is because it is more difficult to predict important pathways in character-level prediction tasks than in word-level tasks due to the higher granularity of the data. Future work may uncover the important pathways with a higher success rate and thus form better ensembles.

\begin{table}[!t]
  \centering
  \caption{\small Self-ensembling results by unbiased SSA on the PCQM4Mv2 dataset, produced with 50 samples for each input graph. Renormalized results are from Table \ref{tab:graph_res}.}
  \label{tab:graph_ens}
  \scalebox{0.77}{
    \newcommand{\cmscol}{\textbf{C}$\downarrow$ / \textbf{M}$\downarrow$ / \textbf{S}$\uparrow$}
\begin{tabular}{l|cc|cc}
    \toprule
      \multicolumn{1}{l|}{}  & \multicolumn{4}{c}{\textbf{dev MAE}$\downarrow$}                                                                          \\
      \multicolumn{1}{l|}{}  & \multicolumn{2}{c|}{\textbf{w/o FT}}                        & \multicolumn{2}{c}{\textbf{+ FT}}                           \\
      \textbf{Model}         & \textbf{Renorm.}              & \textbf{Ensemble}           & \textbf{Renorm.}              & \textbf{Ensemble}           \\ \midrule
      S6-U10                 &          0.0907               &  0.0880                     &              0.0895           &  0.0884                     \\ 
      S6-U20                 &          0.0895               &  \textbs{0.0865}            &              0.0876           &  0.0877                     \\ 
      S6-U30                 &          0.0904               &  \textgd{0.0872}            &              0.0876           &  0.0892                     \\ 
      S6-U40                 &          0.0930               &  0.0893                     &              0.0879           &  0.0923                     \\ 
      S6-U50                 &          0.0964               &  0.0945                     &              0.0908           &  0.1005                     \\ 
    \bottomrule                                                             
\end{tabular}
    }
\end{table}

Self-ensembling can be done for unbiased SSA and regression tasks as well. The results of self-ensembling on the PCQM4Mv2 dataset are presented in Table \ref{tab:graph_ens}. We take an average of 50 sample predictions for each input graph while following the same SSA scheme during inference as during training. We see that the self-ensembling results are better than the renormalized results for all models that have not been fine-tuned. The self-ensembled results are even better than that of renormalized fine-tuned results. This shows that self-ensembling can serve as an alternative to fine-tuning. We believe that the better results are due to the regularization effect of SSA, sampling sub-models that consider sparse and long-range dependencies. These results degrade with fine-tuning because the pathways within these models become less predictable by unbiased SSA after fine-tuning.

\begin{figure}[!t]
  \centering
  \includegraphics[width=1.0\columnwidth]{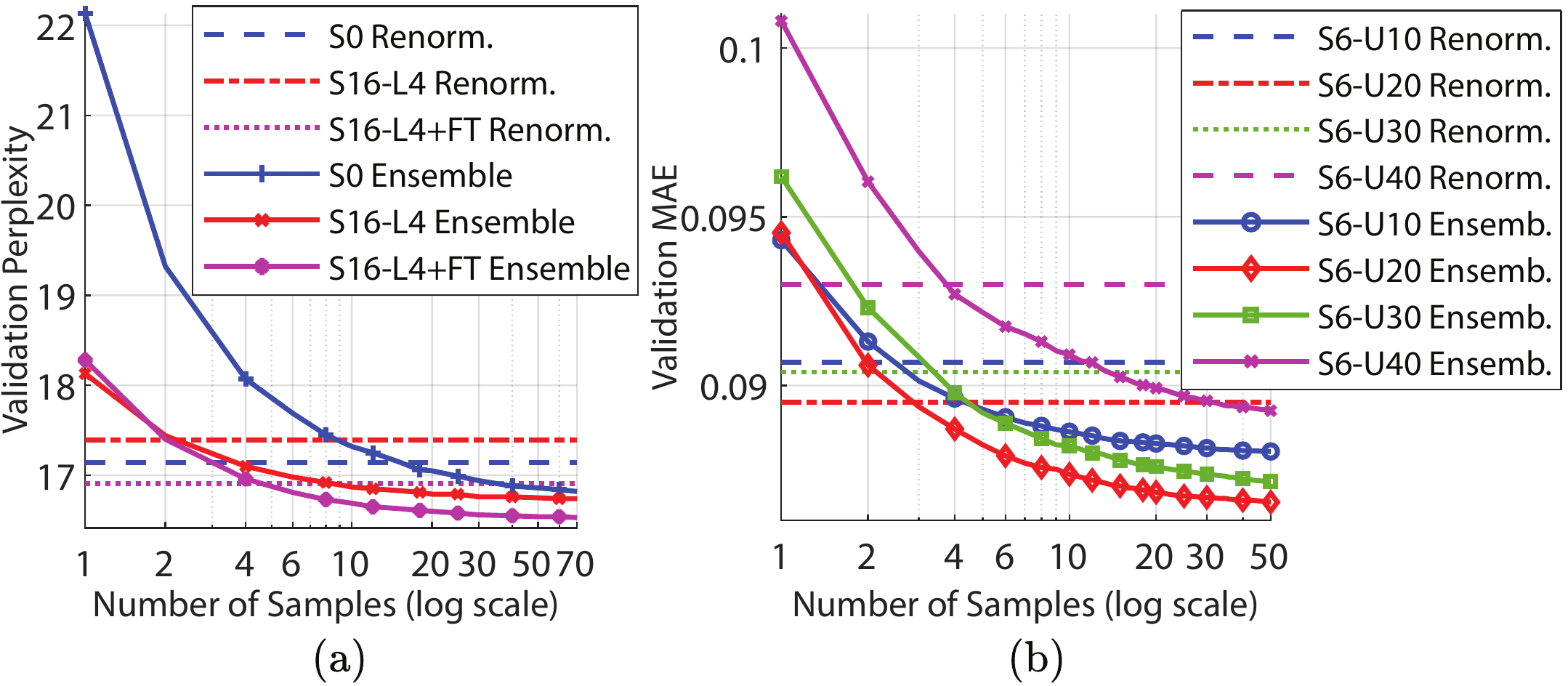}
  \caption{\small Self-ensemble performance for language modeling on (a) WikiText-103 and graph regression on (b) PCQM4Mv2 as a function of the number of samples drawn. Dashed lines show the performance of the renormalized model.}
  \Description{A graph showing the performance of SSA-based self-ensembling as a function of the number of samples drawn.}
  \label{fig:ensemble}
\end{figure}

Fig.~\ref{fig:ensemble} shows how the self-ensembling performance improves with the number of samples drawn, for the language modeling task on WikiText-103, and the graph regression task on PCQM4Mv2, and how they compare against the renormalized results. We see that the self-ensembling performance improves with the number of samples drawn for both tasks. From Fig.~\ref{fig:ensemble} (a) we see that for S0, which was not trained with SSA, we need to draw upwards of 20 samples to improve the results beyond that of renormalization. But for S16-L4 and their fine-tuned counterparts, which were trained with SSA, we need to draw only 2-5 samples to improve the results beyond that of renormalization. Since we are using SSA at inference time, these samples are faster to produce for the sub-models than the full model. This shows that self-ensembling is a practical option for improving the results of a model that was trained with SSA. We believe the important information pathways are more predictably sampled within a model that was trained with SSA, which leads to the result plateauing with fewer samples. However, this rate of improvement also depends on the amount of sparsity applied by SSA. From Fig.~\ref{fig:ensemble} (b) we see that for the graph regression task, we also need to draw only 3-5 samples to improve the results beyond that of renormalization, but S6-U10 which applies only 10\% attention drop plateaus much faster than S6-U50 which applies 50\% drop. This is because variance increases with the amount of sparsity, but this also produces a more diverse set of sub-models, which often leads to better results.

In Fig.~\ref{fig:ensemble}, we observe that even when we draw only a single random sample, the results are not significantly worse than the renormalized results. It is important to note that if the information pathways were \emph{not} independent, randomly selecting a set of pathways to form a sub-model would lead to a drastic drop in performance. This shows that the information pathways are indeed independent, e.g., the presence/absence of a particular pathway does not negatively affect the performance of another pathway. We hypothesize that for a single random sample, the reduction in performance is only due to the reduced strength of the ensemble due to the missing pathways, which is quickly recovered as we draw more samples by covering most of the important pathways. Also, the fact that a few sub-models drawn from a predefined distribution can be as performant as the full model shows that the distribution of the information pathways is predictable.

\section{Conclusion and Future Work}
In this paper, we presented the information pathways hypothesis which states the existence of sparsely connected sub-networks within the transformer called information pathways. A sub-model formed from a random subset of these pathways can be trained at each training step to reduce the cost of training. We introduce an algorithm called SSA which can take advantage of this fact by stochastically sampling only a subset of attention sources and training the important information pathways with a high probability, which not only reduces training cost but also improves generalization. SSA can be applied to any model that uses dense self-attention, and for both generative and discriminative tasks. We showed the effectiveness of SSA for language modeling, image classification, and graph regression tasks. We also showed that SSA can be applied at inference time to form an ensemble of sub-models from the transformer which can further improve the results beyond that of the full model, by making more robust predictions. We used local bias to improve the performance of SSA by sampling the important pathways with a higher probability. 

Our SSA algorithm is simple and easy to implement, but its performance can be further improved by using more sophisticated sampling strategies. The information pathways hypothesis calls for more research into the search for sparsely connected sub-networks within the transformer, and how to better sample them, which could further alleviate the training cost of the transformers while helping them to generalize better using strategies such as attention self-ensembling. We also want to explore the prospect of extending SSA to cross-attention, for tasks such as machine translation.

\begin{acks}
  This work was supported by the Rensselaer-IBM AI Research Collaboration, part of the IBM AI Horizons Network.
\end{acks}

\bibliographystyle{ACM-Reference-Format}
\bibliography{citations/citations, citations/dnn, citations/new}

\appendix

\section{Data and Code Avalability}
\textbf{Data:} All datasets used in this work are publicly available. The dataset sources are listed in Table \ref{tab:datasources}.

\noindent
\textbf{Code:} The code is available at \url{https://github.com/shamim-hussain/ssa}.

\begin{table}[!h]
    \centering
    \caption{Dataset sources.}
    \label{tab:datasources}
    \scalebox{0.83}{
        \begin{tabular}{lc}
    \toprule
    \textbf{Dataset}     & \textbf{Source}                                          \\\midrule
    WikiText-103         & \url{https://huggingface.co/datasets/wikitext}           \\
    Enwik8               & \url{http://mattmahoney.net/dc/textdata.html}            \\
    CIFAR-10             & \url{https://www.cs.toronto.edu/~kriz/cifar.html}        \\
    ImageNet-1k          & \url{https://image-net.org/}                             \\
    PCQM4Mv2             & \url{https://ogb.stanford.edu}                           \\
    \bottomrule                                                                           
\end{tabular}%
        }
\end{table}

\section{Hyperparameters and Training Details}
\subsection{Generative Language Modeling}
For language modeling on both WikiText-103 and Enwik8 datasets, we used the 16-layer transformer decoder of \citet{press2021train} which uses ALiBi relative positional encodings. We used the fairseq toolkit \citep{ott2019fairseq} to perform these experiments. We used an input length of 3072 tokens for WikiText-103. Adaptive input embeddings \citep{baevski2018adaptive} and adaptive softmax \citep{joulin2017efficient} output were used to handle a large vocabulary of size around 260K. For Enwik8, we used a simple vector embedding and a vanilla softmax output layer. We used the same architecture and hyperparameters as \citet{press2021train}, except that we changed the activation function from ReLU to GELU \citep{hendrycks2016gaussian}. For Enwik8, we also add a final Layer Normalization \citep{ba2016layer} layer before the softmax layer. On WikiText-103 we trained for 64,000 steps (16,000 linear learning rate warmup steps, followed by 48,000 steps of cosine decay) with the Adam \citep{kingma2014adam} optimizer, a maximum learning rate of $0.001$ and an increased batch size of 64. For Enwik8, we trained for 10,000 steps (4,000 warmup steps followed by linear decay) with a maximum learning rate of 0.001 and a minimum learning rate of 0.0005. Again we used a batch size of 64 and the Adam optimizer.

We tune the SSA parameter $\sigma$ (in Eq. \ref{eq:locperm}) in different layers for the best validation set results. We express its value as a fraction of the input length so that it is independent of the input length. For WikiText-103 we use a value of $\sigma=0.2$ in the first layer and linearly increase it to $\sigma=0.35$ in the deepest layer. For Enwik8 we start with $\sigma=0.1$ and linearly increase it to $\sigma=0.225$ in the deepest layer.

\subsection{Image Generation and Classification}
For image generation on CIFAR-10, we use a 16-layer transformer with similar architectural and hyperparameter settings as in the previous section, but we use the 2D relative position bias for positional encoding, similar to \citep{liu2021swin}. Instead of using dropout regularization, we use the augmentation techniques described in \citep{jun2020distribution} to achieve better generalization. We use the Adam optimizer with a maximum learning rate of 0.002 and a batch size of 128. We train for 100,000 steps in total, with an initial learning rate warmup of over 4,000 steps, followed by cosine decay. To perform locally biased SSA on this 2D data we divide the image vertically into 4 windows of 8 rows. Locally biased source shuffling is performed in both the horizontal and vertical directions while preserving causality at the window level. We set the SSA parameter $\sigma$ to 0.25 times dimensions (length/width), in both the horizontal and vertical directions, and all layers.

For ImageNet-1K classification, we use the Swin-Tiny model with 12 layers and 28 million parameters and an input resolution of 224x224. We use the same architecture, hyperparameters, augmentation, and training scheme as in \citep{liu2021swin}. To apply locally biased SSA within the 14x14 windows, we further subdivided the windows into 4, 7x7 sub-windows. Locally biased source shuffling was performed in both horizontal and vertical directions, with the value of $\sigma$ as 0.75 times the window size (i.e., 14), but we further ensured that sources were constrained within their own 14x14 (shifted) windows after shuffling (yet they can move beyond the smaller 7x7 sub-windows).

\subsection{Molecular Graph Regression}
For graph regression on the PCQM4Mv2 dataset, we use the Edge-augmented Graph Transformer (EGT) described in \citep{hussain2022global}. EGT uses additional channels to represent and update edge embeddings, which makes it slightly different from the standard transformer. However, we experiment with an ablated variant of EGT called EGT-Simple which reduces the edge representations to relative positional encodings. However, for brevity, we call this model EGT. We experiment on the EGT\textsubscript{small} model with 11 million parameters and 6 layers. We use the same hyperparameters and training scheme as in \citep{hussain2022global}, except, we do not use attention dropout in these models because SSA works as a regularization method.

\begin{table}[!h]
  \centering
  \caption{\small Baseline (S0) training cost for different datasets.}
  \label{tab:base_cost}
  \scalebox{0.77}{
    \begin{tabular}{lcccc}
    \toprule
                         &                                   & \textbf{Compute}    & \textbf{Memory/GPU}  & \textbf{Time/step}   \\
    \textbf{Dataset}     & \textbf{Model}                    & \textbf{(Exa FLOP)} & \textbf{(GB)}        & \textbf{(ms)}        \\\midrule
    WikiText-103         & Transfo. Decoder                  &  7.6                &   21.3               &    492               \\
    Enwik8               & Transfo. Decoder                  &  1.7                &   30.5               &    628               \\
    CIFAR-10             & Transfo. Decoder                  &  23.8               &   29.1               &   1059               \\
    ImageNet-1k          & Swin-Tiny                         &  3.2                &   4.0                &    146               \\
    PCQM4Mv2             & EGT\textsubscript{small}          &  0.5                &   --                 &    --                \\
    \bottomrule                                                                           
\end{tabular}
    }
\end{table}

\section{Baseline Training Cost}
In our experiments, we normalized the training costs with respect to the baseline model S0, the dense attention model without SSA. However, we also report the baseline training costs in terms of absolute values in Table \ref{tab:base_cost} for completeness. Note that for the PCQM4Mv2 dataset, we could not faithfully compute the memory consumption and the training time due to the data loading bottleneck. However, we can still compare the cost of the baseline models with SSA models in terms of compute.


\section{Locally Biased vs Unbiased SSA}
In the results presented in our experiments, we claimed that local bias was an important ingredient in improving the performance of SSA. Here, we directly compare the results for locally biased SSA and unbiased SSA for the same level of sparsity and when they are applied to the same subset of layers.

\begin{table}[!h]
  \centering
  \caption{\small Locally biased vs unbiased SSA results for language modeling tasks on WikiText-103 and Enwik8. Locally biased SSA results are from Table \ref{tab:text_res}.}
  \label{tab:text_lvu}
  \scalebox{0.77}{
    \begin{tabular}{l|cc|cc}
    \toprule
                         &\multicolumn{2}{c|}{\textbf{Wikitext-103}}                                 & \multicolumn{2}{c}{\textbf{Enwik8}}                                     \\
    \textbf{\% Attention}&\multicolumn{2}{c|}{\textbf{dev/test Ppl.} $\downarrow$}                   & \multicolumn{2}{c}{\textbf{dev/test BPB} $\downarrow$}                  \\
    \textbf{Sampled}     &\textbf{Locally Biased}             &\textbf{Unbiased}                     & \textbf{Locally biased}         &\textbf{Unbiased}                      \\ \midrule
    50\% attention       &          17.12 / 17.84             & 17.45 / 18.18                        &           1.052 / 1.028         & 1.114 / 1.087                         \\
    +FT                  &          16.95 / 17.68             & 16.96 / 17.69                        &           1.050 / 1.026         & 1.063 / 1.042                         \\ \midrule
    25\% attention       &          17.39 / 18.13             & 19.33 / 20.15                        &           1.081 / 1.058         & 1.329 / 1.287                         \\
    +FT                  &          16.91 / 17.60             & 17.89 / 18.69                        &           1.052 / 1.029         & 1.100 / 1.075                         \\
    \bottomrule
\end{tabular}%
    }
\end{table}

The results for language modeling are presented in Table \ref{tab:text_lvu} where the same level of sparsity is applied to all layers for both types of SSA (using 2 or 4 window attention for 50\% or 25\% sampling, respectively). We see that unbiased SSA performs slightly worse than locally biased SSA for 50\% sampling of attention. This gap can be reduced with fine-tuning. However, when we only sample 25\% of attention, training is significantly hampered for unbiased SSA, and the results cannot be made comparable to locally biased SSA, even with fine-tuning. This is because unbiased SSA cannot sample the important pathways with a high enough probability for the training to progress gracefully. This shows the necessity of local bias for sampling at high sparsity levels.

\balance
The results for image classification are presented in Table \ref{tab:swin_lvu} where the same level of sparsity (25\% sampled, 75\% dropped) is applied to 10 layers (using 4 sub-windows) or the first 4 layers are excluded. In all cases, we see that locally biased SSA performs better than unbiased SSA, but we do get good results with unbiased SSA when we exclude the first 4 layers. This shows that local bias is important for SSA to work well, but it is less important in deeper layers than in the shallower layers. In deeper layers, the model tends to form long-distance dependencies, which are more predictable by unbiased SSA. This is why we see that unbiased SSA performs better when we exclude the first 4 layers.

\begin{table}[!h]
  \centering
  \caption{\small Locally biased vs unbiased SSA results on the ImageNet-1K image classification task. Locally biased SSA results are presented from Table \ref{tab:img_res}.}
  \label{tab:swin_lvu}
  \scalebox{0.77}{
    \begin{tabular}{l|cc}
    \toprule
    \textbf{\# Layers}   &\multicolumn{2}{c}{\textbf{dev Acc.} $\uparrow$}          \\
    \textbf{Sampled}     &\textbf{Locally Biased}       &\textbf{Unbiased}          \\ \midrule
    10 layers            &          80.56\%             & 80.21\%                   \\
    +FT                  &          81.15\%             & 80.80\%                   \\ \midrule
    6 layers             &          81.23\%             & 81.21\%                   \\
    +FT                  &          81.60\%             & 81.36\%                   \\
    \bottomrule                                                                                                           
\end{tabular}%
    }
\end{table}
\begin{table}[!h]
  \centering
  \caption{\small Additional self-ensembling results for language modeling tasks on WikiText-103, produced with 50 samples per input segment. Renormalized results are from Table \ref{tab:text_res}.}
  \label{tab:x_wiki_ens}
  \scalebox{0.77}{
    \begin{tabular}{lcc}
    \toprule
                    &\multicolumn{2}{c}{\textbf{dev/test Ppl.} $\downarrow$}                                 \\
    \textbf{Model}  &\textbf{Renorm.}                                      &\textbf{Ensemble}                \\ \midrule                                                                                                      
    S16-L6          &          17.49 / 18.30                               & 17.01 / 17.80                   \\
    +FT             &          17.09 / 17.86                               & 16.83 / 17.53                   \\\midrule                                                                                                              
    S12-L8          &          17.94 / 18.69                               & 17.41 / 18.10                   \\
    +FT             &          17.20 / 17.92                               & 17.04 / 17.69                   \\
    \bottomrule                                                                                                           
\end{tabular}
    }
\end{table}

\section{Additional Self-Ensembling Results}
We present additional self-ensembling results for language modeling on WikiText-103 in Table \ref{tab:x_wiki_ens}, for higher levels of sparsity -- with 6 and 8 windows we sample as little as 16.7\% and 12.5\% attention, respectively. We see similar results as presented in the main section with self-ensembling significantly improving over their renormalization counterparts. This shows that self-ensembling can improve performance even with a high level of sparsity.

\end{document}